# SkinSAM: Empowering Skin Cancer Segmentation with Segment Anything Model

Mingzhe Hu[a], Yuheng Li[b] and Xiaofeng Yang[a,b,c*]
[a]Department of Computer Science and Informatics, Emory University, GA, Atlanta, USA
[b]Department of Biomedical Engineering, Emory University, GA, Atlanta, USA
[c]Department of Radiation Oncology, Winship Cancer Institute, School of Medicine, Emory University, GA, Atlanta, USA
*Email: xiaofeng.yang@emory.edu

## Abstract

Skin cancer is a prevalent and potentially fatal disease that requires accurate and efficient diagnosis and treatment. Although manual tracing is the current standard in clinics, automated tools are desired to reduce human labor and improve accuracy. However, developing such tools is challenging due to the highly variable appearance of skin cancers and complex objects in the background. In this paper, we present SkinSAM, a fine-tuned model based on the Segment Anything Model that showed outstanding segmentation performance. The models are validated on HAM10000 dataset which includes 10015 dermatoscopic images. While larger models (ViT_L, ViT_H) performed better than the smaller one (ViT_b), the finetuned model (ViT_b_finetuned) exhibited the greatest improvement, with a Mean pixel accuracy of 0.945, Mean dice score of 0.8879, and Mean IoU score of 0.7843. Among the lesion types, vascular lesions showed the best segmentation results. Our research demonstrates the great potential of adapting SAM to medical image segmentation tasks.

## 1. Introduction

A skin cancer is an abnormal change in the skin's texture, appearance, or color and threatens the health of millions of people worldwide[1]. According to the American Cancer Society, skin cancer is one of the most common types of cancer, with over 5 million patients[2] in the United States. If malignant skin cancers go untreated, they can quickly spread across the body and even threaten patients' lives. So early screening and treatment of skin cancers is crucial to reduce the patient's suffering and improve the outcome. Skin cancer segmentation is a task to identify and delineate the locations and boundaries of the cancers. Fast and accurate skin cancer segmentation can ensure disease monitoring, diagnosis, treatment planning, and prognosis effectiveness and efficiency. While manual tracing of the cancers is still the standard way in clinics, automated tools are desirable for reducing human-labors and professional expertise and low inter-observer variability[3]. However, developing such reliable tools is difficult due to the highly different cancer appearance in shape, size, color texture. Besides, the images also include other complex objects in the background, like sweat glands, blood vessels, and hair.

In the past decade, the fast development of models based on R-CNN[4], U-Net[5], vision transformers[6], and mlp-mixer[7] structures has revolutionized the domain of medical image segmentation and showed unprecedented performance[8-11] compared to the traditional segmentation algorithms. Despite the success of supervised or unsupervised approaches built with these models, limitations still exist. Supervised learning methods usually need sufficient ground truth labels during the training process. The large and high-quality annotated dataset is scarce for medical images due to the requirement of high human labor and expert knowledge during the labeling process. Though some unsupervised methods exist, lacking the ground truth of the segmentation masks, the accuracy and reliability of these methods are not guaranteed compared with the supervised ones.

The advent of zero-shot learning[12] provides a promising alternative to address the challenges of traditional supervised and unsupervised machine learning models. It enables a model to segment an object type it has never encountered. In zero-shot learning, a model is trained on a source domain to perform segmentation and build prior knowledge. The segmentation of the new type of object is achieved by mapping the visual features of an object to a semantic space to build up the relationship between the seen object type and the unseen type.

Recently, a zero-shot learning method called the Segment Anything Model (SAM)[13] achieved state-of-the-art performance on several benchmark datasets and has provoked the interest of many researchers in the computer vision domain. Besides the simplicity of the pipeline it provides for domain adaptation, its success, and popularity are due to several reasons. First of all, SAM is very robust. It was trained on the SA-1B dataset[13], which contains over 11 million images and 1 billion masks, enabling it to understand the visual concepts deeply. Second, to ensure the quality of new segmentation tasks, users can show their intent by providing prompts at inference time. Finally, the model was built with advanced techniques like multi-scale feature fusion and attention mechanism, which enabled the excellent capture of local and global features in the image.

In this research, we adapted the SAM model for skin cancer segmentation. We discussed the impact of prompts on the segmentation results, as well as compared the performance of three different pre-trained model structures (ViT-b, ViT-l, ViT-H). We demonstrated that fine-tuning the pre-trained models can lead to further improvements in segmentation performance. By the end of this paper, the readers should clearly understand the limitations and future perspectives of using SAM for skin cancer segmentation.

## 2. Method
### 2.1 HAM10000 Dataset

The HAM10000[14] dataset collects 10,015 dermatoscopic images representing a comprehensive range of pigmented skin lesions. This dataset includes a diverse set of diagnostic categories, including actinic keratoses and intraepithelial carcinoma (AKIEC), basal cell carcinoma (BCC), benign keratosis-like lesions (BKL), dermatofibroma (DF), melanoma (MEL), melanocytic nevi (NV), and vascular lesions (VASC). In addition to the images, the binary segmentation masks[15] for each image are also provided, created by a dermatologist to identify the lesion areas. The masks were initially generated using a fully connected neural network (FCN) lesion segmentation model and then manually corrected or redrawn using the free-hand selection tool in FIJI[16]. Figure 1 displays example images and masks for each of the seven lesion types in the dataset in this study.

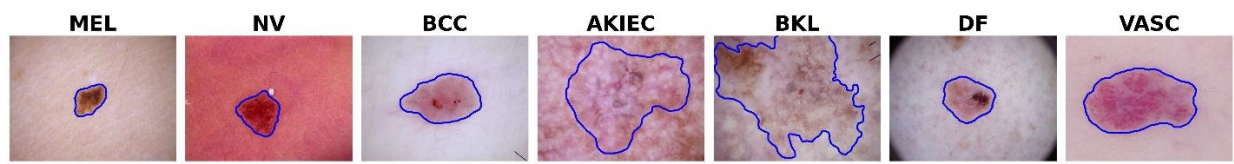

Figure 1: Example images and masks for each of the seven lesion types in the dataset. Each image is accompanied by its corresponding mask, with the lesion contour outlined in blue. The seven lesion types represented are Melanoma (MEL), Melanocytic nevus (NV), Basal cell carcinoma (BCC), Actinic keratosis / Bowen's disease (AKIEC), Benign keratosis-like lesions (BKL), Dermatofibroma (DF), and Vascular lesions (VASC).

### 2.2 Segment Anything (SAM)

The backbone structure of the SAM model is based on a pre-trained Vision Transformer (ViT) that has been minimally adapted to process high-resolution inputs. Specifically, the image encoder component of the SAM model uses an maksed auto encoder (MAE)[17] pre-trained ViT[6] that has been modified to handle input images of up to 1024x1024 resolution. The ViT backbone consists of a series of self-attention layers that allow the model to capture long-range dependencies between different parts of the input image. This is followed by a set of feedforward layers that transform the output of the self-attention layers into a set of feature maps that are used by the mask decoder component to generate segmentation masks. Overall, this backbone structure allows the SAM model to efficiently process high-resolution images while capturing fine-grained details and long-range dependencies in the input.

SAM provides users the option to use prompts to indicate their segmentation intents. According to the original paper, the prompts can be in various forms, such as a natural language description, a sketch, or an example image. Here, for simplicity, we adopted the sketch method to indicate the foreground we would segment using points and bounding boxes. Since most of the lesion in this research is contiguous, we only use one set of prompt for each image. Subsequently, the prompt encoder component of the SAM model encodes the prompt and produces a fixed-length embedding that captures the semantic meaning of the prompt. This embedding is then combined with the output of the image encoder component to generate a set of feature maps that the mask decoder component uses to generate segmentation masks. We take the mask with the highest probability as our final mask. Figure 2 shows the output masks of a sample image.

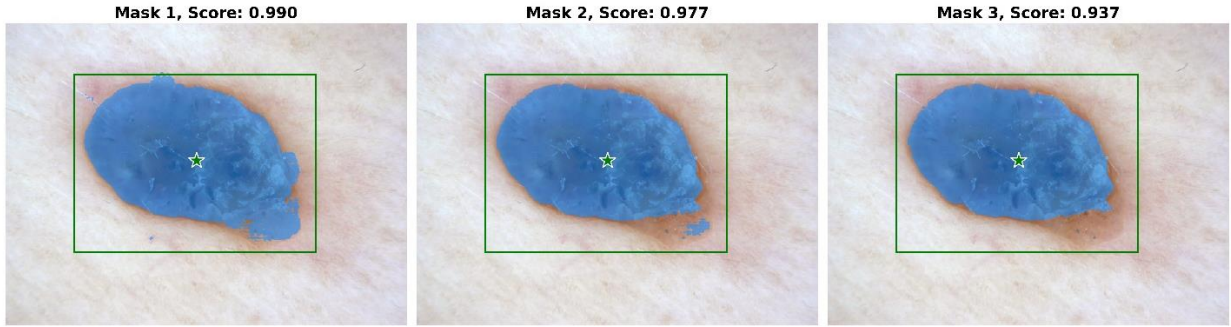

Figure 2: Three different masks generated by the predictor on the sample image. The masks are plotted in blue while the prompts (boxes and points) are plotted in green. The mask with the highest probability was chosen as the final segmentation result.

2.3 Prompts Simulation

However, when validating on a large dataset, it is virtually impossible to manually sketch the prompts. Therefore, to simulate the prompt generation process, we first generate "ground prompts" based on the ground segmentation masks and then introduce some randomness. The pseudocode for this process is presented below:

for each image in the dataset:
    load the image and corresponding ground truth mask
    randomly select a point within the mask
    define the size of the bounding box to be the dimensions of the mask plus 20 pixels on each side
    randomly shift the bounding box horizontally and vertically by up to 30 pixels in either direction
    randomly scale the bounding box by up to 10%
    overlay the bounding box and point on the image as prompts

2.4 Finetuning the Model

We divided the dataset into training (80%) and validation (20%) through random sampling. The image encoder employed the pre-trained ViT-Base model. All image embeddings were computed offline by supplying the normalized images to the encoder, which then resized them to dimensions of $3 \times 1024 \times 1024$. We derived the bounding box prompt from the ground-truth mask using the method we showed in the previous section.

Our chosen loss function combined Dice loss and cross-entropy loss without any weighting, a strategy that has demonstrated effectiveness in a wide array of segmentation tasks. To optimize the network, we used the Adam optimizer and set the initial learning rate to 1e-5. To decide when to stop training, we monitored the performance of the validation set. We ceased training when the model's performance plateaued, indicating that additional training would not yield significant improvements and could lead to overfitting.

The PC we used to finetune the model has an x86_64 architecture with 12 CPUs, 6 cores per socket and 2 sockets, for a total of 24 threads. The CPU model is an Intel Xeon E5-2603 v4 @ 1.70GHz with 30MB L3 cache. The system has 46-bit physical and 48-bit virtual address sizes, and supports virtualization via VT-x. The GPU is a Tesla V100-PCIE with 2 cards, each with a memory capacity of 32GB.

2.5 Evaluation Metrics

In order to evaluate the performance of our segmentation model, we used several metrics, including pixel accuracies, Intersection over Union (IOU), and Dice Score. Pixel accuracies measure the percentage of pixels that the model correctly classifies. IOU measures the overlap between the predicted and ground truth masks by dividing the intersection of the masks by the union of the masks. Dice Score is another measure of overlap between the predicted and ground truth masks, which is calculated by computing the harmonic mean of the precision and recall scores. The definition of different metrics is shown as table 1.

Table 1. Evaluation metrics we used for this study and their definitions.

| Metric | Formula |
|---|---|
| Pixel Accuracy | Number of correctly predicted pixels / Total number of pixels |
| IOU (Jaccard) | Intersection over Union = Intersection / (Prediction + Ground Truth - Intersection) |
| Dice Score | 2 * Intersection / (Prediction + Ground Truth) |

3. **Result**

The overall performance of our models with prompts is quite good. Here, we first visualize some segmentation results in figure 3.

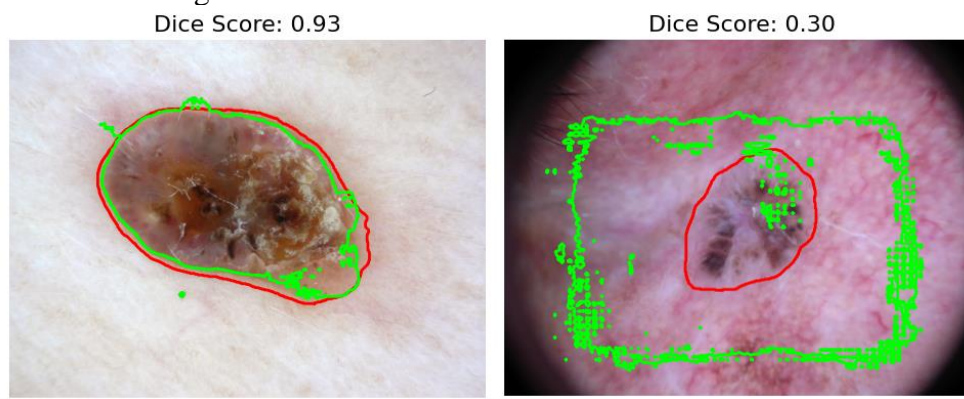

Figure 3: Sample skin lesion images and their corresponding ground truth and prediction masks with contour overlaying. The red contour represents the ground truth, while the green contour represents the prediction. The image on the left shows a successful prediction, whereas the image on the right shows an unsuccessful prediction.

3.1 Prompt interaction vs. No Prompt

One of the main features of the SAM system is prompt interaction. We compared the performance of skin cancer segmentation using ViT-b model with and without prompts. As shown in Figure 4, the segmentation results with prompt are highly satisfactory. However, without prompt, the model is unable to comprehend the user's intention, resulting in a prediction mask with the highest probability that still falls short of the ground truth. The mean pixel accuracy, mean dice and mean IoU score with prompt were 0.8675, 0.8125, and 0.6952, respectively. Conversely, the mean pixel accuracy, mean dice, and mean IoU score without prompt were 0.2087, 0.2400, and 0.1628, respectively.

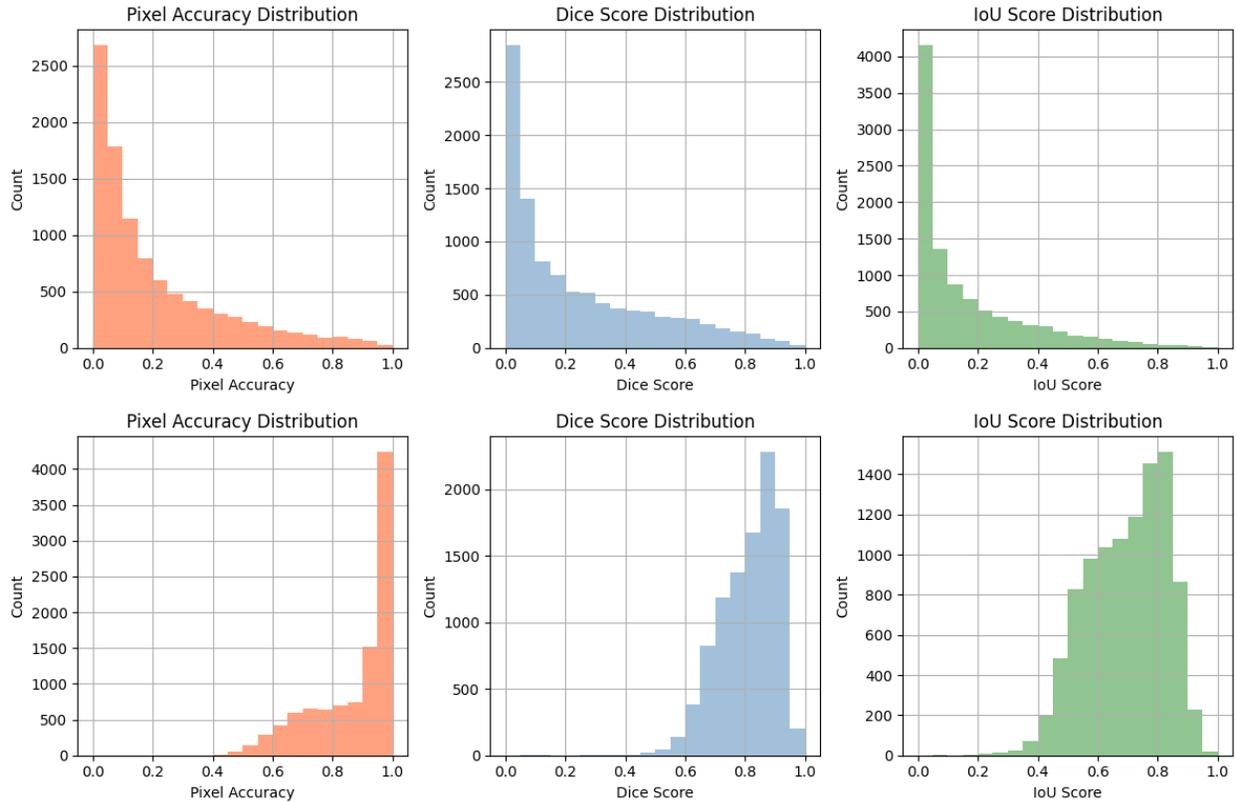

Figure 4: Histograms of segmentation scores with and without prompt. Each column represents a different evaluation score. The first row represents the results without prompt, where most of the results are clustered on the left side of the histogram, indicating poor segmentation performance. On the other hand, the second row represents the results with prompt, where significant improvement is observed in segmentation performance.

3.2 Comparison of Different Lesions

Figure 5 shows the histograms of segmentation scores for each type of skin lesion using ViT-b. Table 1 shows the detailed results. Based on the results, we can see that the highest mean scores are for the lesion types VASC and NV in all three metrics (Dice Score, IoU Score, and Accuracy). The lowest mean scores are for the lesion types AKIEC and BCC. The variance in the scores is relatively low across all lesion types and metrics, with the highest variance being 0.02. This suggests that the model performs consistently across different lesion types and metrics. Overall, the results suggest that the model performs well in accurately detecting and segmenting different lesion types, with some variation in performance depending on the specific lesion type and metric being used.

Table 1: The mean and variance of dice score, IoU score, and accuracy for each lesion type in the format of mean(variance).

| Lesion Type | Dice Score | IoU Score | Accuracy |
| --- | --- | --- | --- |
| MEL | 0.81(0.01) | 0.70(0.02) | 0.84(0.02) |
| VASC | 0.86(0.01) | 0.77(0.02) | 0.96(0.00) |
| NV | 0.82(0.01) | 0.71(0.02) | 0.89(0.01) |
| BKL | 0.80(0.01) | 0.68(0.02) | 0.83(0.02) |
| BCC | 0.74(0.01) | 0.60(0.02) | 0.82(0.02) |
| AKIEC | 0.74(0.01) | 0.60(0.01) | 0.73(0.02) |
| DF | 0.76(0.01) | 0.62(0.01) | 0.82(0.01) |

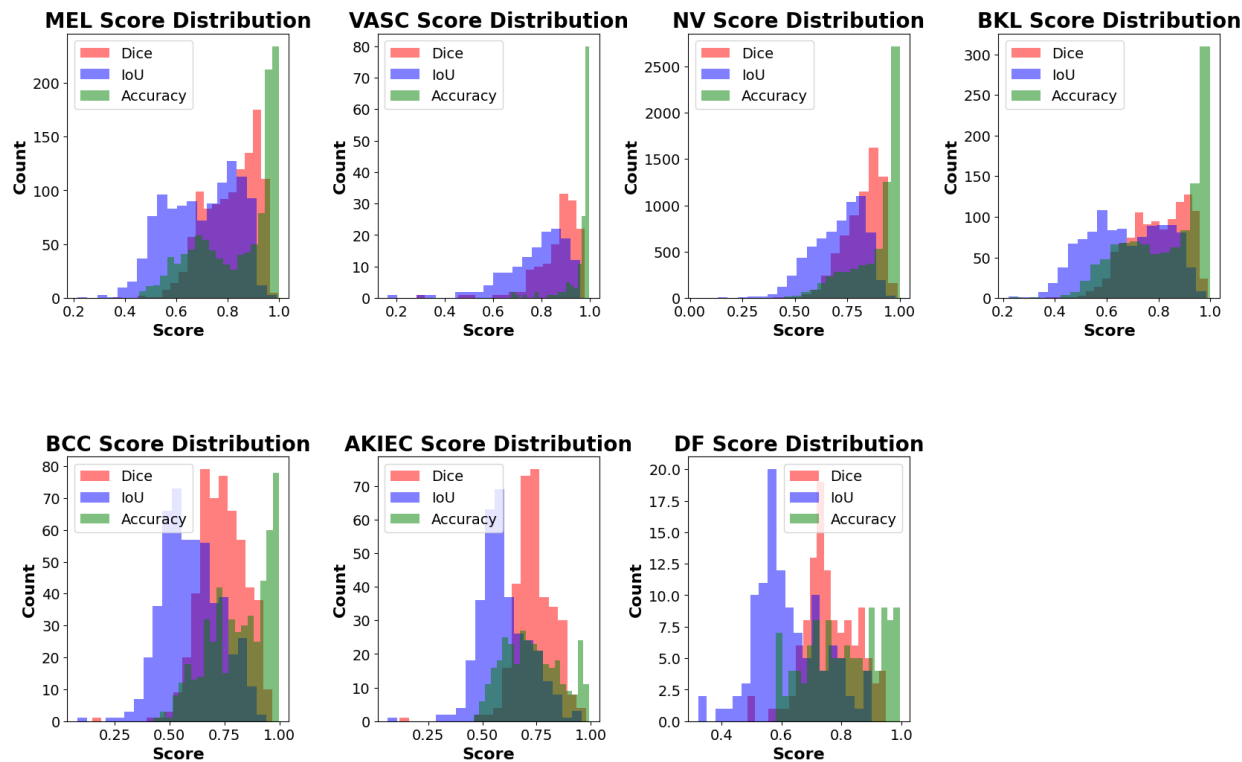

Figure 5: The histograms of segmentation scores for each type of skin lesion using ViT-b. We can observe that VASC has the best score distribution.

3.3 Comparison of Different Models

Three variants of the Vision Transformer (ViT) architecture are used in the SAM model - ViT_b, ViT_l, and ViT_h. These variants differ in the number of layers and hidden units used in the self-attention and feedforward layers of the ViT backbone. Specifically, ViT_b has 91 million parameters, ViT_l has 208 million parameters, and ViT_h has 636 million parameters. The higher number of layers and hidden units in ViT_l and ViT_h allows them to capture more complex patterns in the input image but also requires more computational resources to train and evaluate. In this section, we compare their performance with the finetuned ViT_b model as shown in figure 6.

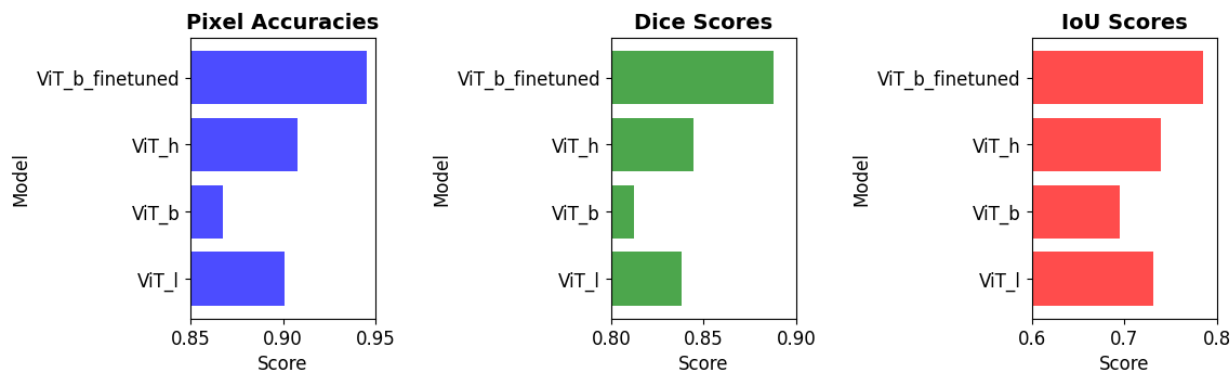

Figure 6: A comparison of the performance of four different ViT models for skin lesion segmentation. The metrics evaluated are pixel accuracy, Dice score, and IoU score. Overall, the results suggest that the ViT_b_finetuned model outperforms the other models in all three metrics. The results of the skin lesion segmentation using ViT-based models show that the model with the highest mean pixel accuracy was ViT_b_finetuned with a score of 0.945, followed by ViT_h with a score of 0.9081, ViT_l with a score of 0.9009, and ViT_b with a score of 0.8675. In terms of the mean dice score, ViT_b_finetuned again had the highest score with 0.8879, followed by ViT_h with 0.8443, ViT_l with 0.8381, and ViT_b with 0.8125. Finally, for the mean IoU score, ViT_b_finetuned had the highest score of 0.7843, followed by ViT_h with 0.7389, ViT_l with 0.7312, and ViT_b with 0.6952. These results suggest that fine-tuning the ViT_b model can lead to significant improvements in performance in all three metrics.

In terms of performance, we can see that the best results are obtained using the ViT_b_finetuned model, with the highest mean scores in all three metrics (pixel accuracy, dice score, and IoU score). The ViT_h model also performs well, with slightly lower scores than the fine-tuned model. The ViT_l model has the lowest scores overall, while the ViT_b model has lower scores compared to the other models but still performs relatively well. To calculate the percentage improvements of ViT_b_finetuned over ViT_b, we used the formula: Percentage Improvement = ((New Value - Old Value) / Old Value) * 100. Using this formula, we found that ViT_b_finetuned improved the pixel accuracy by 8.92%, the dice score by 9.28%, and the IoU score by 12.83% compared to ViT_b. The greatest improvement was observed in the IoU score. These results suggest that larger ViT models may be better suited for this task, and finetuning can improve the model's performance.

4. Discussion

In this study, we attempted to use the zero-shot capability of the SAM system to perform skin cancer segmentation. The results showed that state-of-the-art segmentation performance can be achieved with prompt guidance. However, some issues still need to be explored.

In our research, the results without prompts showed significantly lower performance in segmentation compared to the results with prompts. The potential problems causing this difference include the lack of guidance for the model, which makes it challenging to identify the correct region of interest (ROI) in the image. Additionally, skin cancer images can be complex and contain various objects in the background, making it difficult for the model to distinguish between the lesion and other objects without prompts. The variability in lesion appearance further complicates the task. The SAM model's reliance on prior knowledge from its training dataset may also be a limitation in zero-shot learning without prompts, as it might not establish a clear relationship between unseen object types, such as skin cancer lesions, and the seen object types. To address these problems, it is crucial to use prompts in the SAM model for skin cancer segmentation, guiding the model towards the desired ROI and helping it understand the user's intent.

The use of 2D images in our research presents a limitation when it comes to expanding the methodology to other modalities that involve 3D inputs. While 2D images have been helpful in the segmentation of skin cancer lesions, they only provide a surface representation of the object and may not capture the complete structural information, which can be crucial in some applications. Expanding the methodology to 3D inputs comes with several challenges. First, there is a significant increase in data complexity when dealing with 3D images compared to 2D images. This added complexity requires more sophisticated models and processing techniques to handle the increased dimensionality, which in turn necessitates greater computational resources and longer processing times.

Another point worth mentioning is that we only finetuned the smallest ViT_b model in this study. Although the larger models showed better validation performance than ViT_b, we did not have sufficient time and computation resources to finetune them in this study. However, further finetuning of these larger models is a promising avenue for improving the performance of the skin lesion segmentation task.

SAM being a versatile and adaptable approach, offers numerous possibilities beyond direct use for segmentation tasks. It can assist other segmentation models by providing valuable contextual and spatial information, enhancing their performance. Additionally, SAM can be utilized for generating ground truth labels, expediting the process of creating annotated datasets, and potentially producing more accurate and consistent annotations. By extending its applications, SAM can contribute significantly to the development and improvement of computer vision systems.

## 5. Conclusion

Our paper presents a validation and finetuning of the SAM model for skin cancer segmentation, demonstrating its excellent and robust performance across various skin lesion types. While larger models showed better results, the finetuned model exhibited the most significant improvements. In the future, we aim to adapt SAM for other medical imaging modalities and enable 3D segmentation. Furthermore, we plan to finetune larger models, such as ViT_h, to seek further improvements. We encourage other researchers to build on this research and explore the potential of SAM in the field of medical imaging. This study serves as a launchpad for further exploration of SAM's capabilities and possibilities for improving the accuracy and efficiency of medical image segmentation.